\documentclass[11pt]{article}

\usepackage[preprint]{acl}
\usepackage{enumitem}
\usepackage{times}
\usepackage{latexsym}
\usepackage{multirow}
\usepackage[T1]{fontenc}
\usepackage[utf8]{inputenc}
\usepackage{microtype}
\usepackage{appendix}
\usepackage{inconsolata}
\usepackage{graphicx}
\usepackage{amsmath}
\usepackage{amssymb}
\usepackage{booktabs}
\usepackage{tipa}
\usepackage{float}
\usepackage{listings}
\usepackage{xcolor}
\usepackage{caption}
\usepackage{ragged2e}
\usepackage{tabularx} 

\title{Prior over Evidence: Stereotype-Driven Diagnosis in LLM-Based L2 Pronunciation Feedback}

\author{
  \mdseries\normalsize 
  \begin{tabular}{cc}
    Rong Wang & Kun Sun \\
    \small University of Tuebingen, Tuebingen, Germany & \small Tongji University, Shanghai, China \\
    \small \texttt{rong.wang@uni-tuebingen.de} & \small \texttt{kunsun@tongji.edu.cn}
  \end{tabular}
}
\begin{document}
\maketitle

\begin{abstract}
Large language models are increasingly deployed for written pronunciation feedback in second-language (L2) English learning, under the assumption that their diagnoses are grounded in the supplied speech evidence rather than in priors from pretraining. We test this assumption on $1{,}800$ L2-Arctic utterances ($300$ per L1) spanning six L1 backgrounds, three audio-capable LLMs (Gemini 3.0 Flash, GPT-4o, and Qwen 3.5 Omni-plus), four pronunciation dimensions, and five evidence conditions ranging from a text-only baseline to numeric acoustic features and raw audio. Each (utterance $\times$ model $\times$ condition $\times$ dimension) cell is scored on three metrics: Rating Accuracy (RA) against gold labels, Evidence Coherence (EC) assessing internal consistency without ground truth, and Grounded Correctness (GC) evaluated against gold evidence. Three findings hold across models. First, rating accuracy and grounded reasoning decouple: $39.6\%$ of judged cells contain internally coherent reasoning that supports a wrong rating, against only $15.8\%$ where the reasoning supports a correct rating. Second, phoneme-level feedback collapses onto a fixed inventory of L2-English difficulty phones (/\textipa{T}/, /\textipa{D}/, /\textipa{\*r}/, /\textipa{v}/) that recurs across all six L1 backgrounds and all evidence conditions. Third, acoustic evidence improves the rating only when the supplied feature directly probes the target dimension: textualised F0 range lifts pitch-variation grounding from ($0.18$--$0.19$) to ($0.45$--$0.62$) across all three models, while stress and phoneme correctness, which require target-to-realisation alignment, remain ungrounded. The same audio waveform without textualised F0 values does not reproduce the lift. It is concluded that current general-purpose LLMs are more reliable as verbalisers of externally computed pronunciation evidence than as standalone diagnostic engines.
\end{abstract}

\section{Introduction}
\label{sec:introduction}

Computer-assisted pronunciation training (CAPT) systems increasingly delegate the written-feedback step to large language models (LLMs)~\cite{jeon2024systematic, zhong2024llmfeedback, fu2024multimodalpa}. The promise is concrete. Instead of returning a single goodness-of-pronunciation score~\cite{witt2000phone} that a learner cannot act on, the model can read aligned phonemes, integrate acoustic measurements, and produce human-readable diagnoses~\cite{li2017mispronunciation, wang2025gpt4opa}.
The broader SpeechLM literature has begun to outline a roadmap toward “superhuman” speech understanding, in which LLMs are expected not only to process raw audio but also to reason over its semantic and paralinguistic content~\cite{bu2024sagi, cui2024speechlm_survey}. These works treat this capability as operational and assume the LLM grounds its feedback in the supplied evidence rather than in priors stored from pretraining text. 

This assumption is rarely tested directly. A grounded model and a prior-driven model are externally indistinguishable: both produce confident, fluent prose. The operative difference is whether the feedback varies with the speaker's actual production. A grounded model adapts to the utterance; a prior-driven model issues stereotyped advice, such as warnings about /\textipa{T}/ or /\textipa{\*r}/, regardless of what the speaker produced. Misdirected feedback is more costly than no feedback because it consumes learner attention that would otherwise go to the real error~\cite{jeon2024systematic}.

We address three primary research questions:
\begin{itemize}[leftmargin=*, itemsep=2pt, topsep=2pt]
    \item \textbf{RQ1.} Does providing structured evidence (IPA, acoustic features, or raw audio) improve per-dimension rating accuracy over a baseline?
    \item \textbf{RQ2.} When a model generates structured explanations alongside its rating, does this evidence genuinely justify the diagnosis when validated against ground truth?
    \item \textbf{RQ3.} To what extent are diagnostic errors driven by speaker demographic priors versus L1-independent pedagogical stereotypes?
\end{itemize}

We evaluate three audio-capable LLMs on $1{,}800$ L2-Arctic utterances~\cite{zhao2018l2arctic} across five evidence conditions, reporting Rating Accuracy (RA), Evidence Coherence (EC), and Grounded Correctness (GC) per (model $\times$ condition $\times$ dimension) cell; each response contains both a rating and typed evidence, letting us separate label correctness from explanation grounding.
Our contributions are threefold: (i) an evaluation framework (RA, EC, GC) decoupling label correctness from explanation grounding across five evidence conditions; (ii) empirical proof that $39.6\%$ of evaluated instances exhibit internally coherent but factually incorrect diagnoses driven by a fixed L2-stereotype inventory; and (iii) a per-dimension analysis investigating whether and when explicit acoustic evidence can successfully counteract these priors.

\section{Related Work}
\label{sec:related}

\paragraph{LLM-based pronunciation feedback.}
Classical CAPT pipelines rely on goodness-of-pronunciation scores derived from forced-alignment posteriors~\cite{witt2000phone}, later extended by neural mispronunciation detection and diagnosis (MDD) models that produce per-phone error labels~\cite{li2017mispronunciation, yan2023peppanet}. Recent work couples these signal-level systems with LLMs along three lines. The first uses an LLM to generate articulatory-level explanations on top of MDD outputs~\cite{zhong2024llmfeedback}; the second prompts a multimodal LLM directly for pronunciation scores~\cite{fu2024multimodalpa, wang2025gpt4opa}; the third paraphrases acoustic features into textual prosodic descriptions before passing them to the LLM~\cite{chen2025textpa, qian2025prosodylm}. Our \texttt{text+acoustic} condition belongs to the third family: we supply F0 minimum, F0 maximum, duration, and intensity range as numeric text alongside the IPA transcript. We replicate the textual-evidence advantage on pitch variation reported by~\citet{chen2025textpa} but show that the same advantage does not generalise to stress or phoneme correctness, and that supplying the audio waveform alone does not reproduce the gain on pitch variation, where the textual form does. None of these prior systems separate rating accuracy from explanation grounding and our three-metric framework is designed to fill that gap.


\paragraph{Faithfulness and grounding.}
Faithfulness of generated explanations to model decision processes is a long-standing concern in NLP~\cite{jacovi2020faithfully, maynez2020faithfulness, atanasova2023faithfulness}. Reference-match metrics such as FActScore~\cite{min2023factscore} and AlignScore~\cite{zha2023alignscore} verify claims against supplied sources but assume the source is ground truth. \citet{turpin2023unfaithful} show that chain-of-thought rationales can systematically misrepresent the factors driving a prediction. Our Grounded Correctness metric extends this concern to structured speech feedback: we evaluate the rating, the cited evidence, and the reason jointly against external gold annotations, rather than against any single reference. The $39.6\%$ confabulation rate we observe is the speech-feedback analogue of the unfaithful-CoT finding.

\paragraph{Parametric priors and demographic bias.}
LLMs and speech systems both carry priors that can override input evidence. Question-answering studies show that models often fall back on parametric knowledge when the supplied context is weak~\cite{petroni2019language, mallen2023llms, tao2024contextparametric, kassner2021negated}. In speech technology, automatic speech recognition exhibits higher error rates for non-native and minority-dialect speakers~\cite{koenecke2020racial}. We observe both effects. When the prompt lacks the acoustic feature that probes the target dimension, the model falls back on demographic labels or on a stored ``L2 English problems'' inventory. The phoneme stereotype we document is one such inventory: /\textipa{T}/, /\textipa{D}/, /\textipa{\*r}/, and /\textipa{v}/ dominate the over-claimed phones across all six L1 backgrounds we test, despite the contrastive-analysis tradition predicting L1-specific substitution patterns~\cite{lado1957linguistics, eckman1977markedness, swan2001learner}.


\section{Methodology}
\label{sec:methodology}

\begin{figure*}[t]
    \centering
   \includegraphics[width=0.95\textwidth, height=0.3\textheight]{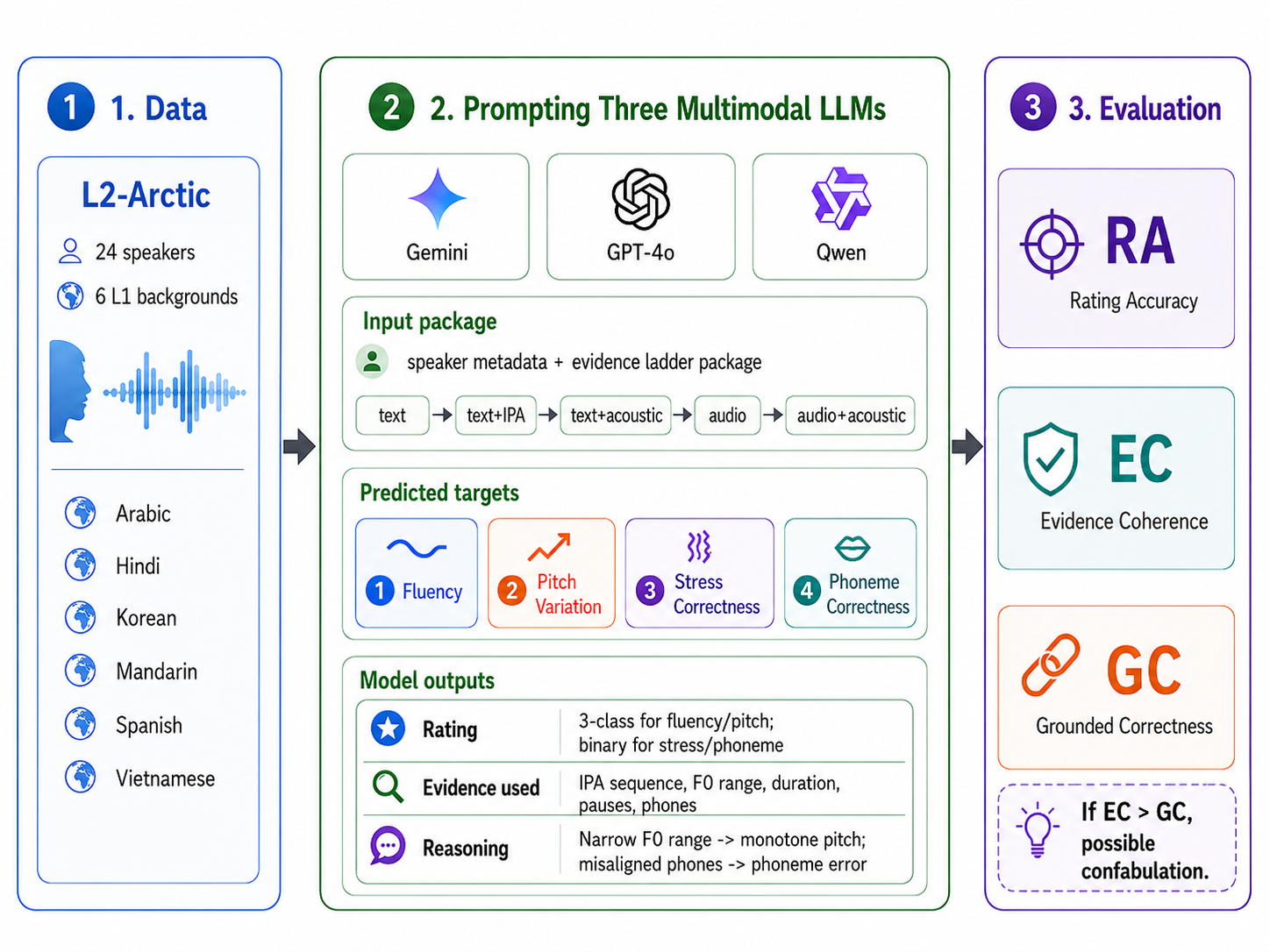}
\caption{\textbf{Evaluation framework.} L2-Arctic utterances from six L1 backgrounds are passed to three multimodal LLMs under five evidence conditions. Each response contains a rating and typed evidence for four pronunciation dimensions. Finally, \textbf{Rating Accuracy (RA)}, \textbf{Evidence Coherence (EC)}, and \textbf{Grounded Correctness (GC)} are evaluated for each model-condition-dimension cell.}
    \label{fig:framework}
\end{figure*}

The design separates three components that pronunciation-feedback studies often conflate: the speech material, the evidence supplied to the model, and the metrics used to judge correctness and grounding. Figure~\ref{fig:framework} gives an overview.

\subsection{Dataset and gold targets}
\label{sec:dataset}

We use L2-Arctic~\cite{zhao2018l2arctic}, a read-speech corpus of 24 non-native speakers across six L1 backgrounds (Arabic, Hindi, Korean, Mandarin, Spanish, Vietnamese) with human-verified TextGrid annotations for phone-level errors and lexical stress. From the human-verified portion, we sample $300$ utterances per L1 ($1{,}800$ utterances total) by speaker-balanced round-robin sampling: within each L1 the procedure shuffles the available speakers, then iteratively draws one utterance from each speaker until the target is reached. Speaker balance prevents a small number of speakers from dominating the acoustic or error distribution.

We define a gold rating and gold evidence target along four dimensions. For \texttt{fluency} and \texttt{pitch\_variation} (three-class: \texttt{slow}/\texttt{normal}/\texttt{fast}; \texttt{monotone}/\texttt{normal}/\texttt{varied}), reference labels come from words per second and from F0 range, each binned by within-corpus z-score at $\pm 0.5$ SD. For \texttt{stress\_correctness} and \texttt{phoneme\_correctness} (binary), labels come from the L2-Arctic TextGrids: stress is positive if any stress-bearing vowel in the utterance differs from canonical stress; phoneme is positive if any phone is annotated as substitution, deletion, or addition. Gold evidence for the binary dimensions is the validated set of errored stressed vowels or errored phones. On the scorable subset the gold positive rate is $3.8\%$ for stress and $96.3\%$ for phoneme; the corresponding always-positive Detection-F1 baselines are $0.07$ and $0.98$, which motivates the auxiliary grounding metrics in §\ref{sec:metrics}.


\subsection{Evidence conditions}
\label{sec:conditions}

We evaluate five conditions that vary the prompt's evidence package while holding the response format fixed (Table~\ref{tab:evidence_conditions}). \texttt{text-only} supplies the target sentence, speaker L1, and gender; \texttt{text+ipa} adds the canonical IPA transcript; \texttt{text+acoustic} adds the numeric acoustic features as text; \texttt{audio-only} supplies the raw waveform alongside IPA but withholds the numeric features; \texttt{audio+acoustic} provides the waveform and the numeric features together. Acoustic fields are supplied as primitive measurements (duration, F0 minimum, F0 maximum, intensity range) rather than as pre-computed diagnostic labels (e.g., words per second, F0 range), so the model must perform a derivation to use them. This design choice differs from~\citet{chen2025textpa}, who paraphrase the same measurements into textual prosodic descriptions, and prevents the grounded conditions from becoming label-copying tasks~\cite{tao2024contextparametric}.

\begin{table}[t]
\centering
\footnotesize

    \caption{Five Evidence Conditions}
    \label{tab:evidence_conditions}
    
    \setlength{\tabcolsep}{3pt} 
    \begin{tabular}{lccccc}
    \toprule
    Condition & Sent. & L1/G & IPA & Acous. & Audio \\
    \midrule
    \texttt{text-only}     & \checkmark & \checkmark & --         & --         & --         \\
    \texttt{text+ipa}      & \checkmark & \checkmark & \checkmark & --         & --         \\
    \texttt{text+acoustic} & \checkmark & \checkmark & \checkmark & \checkmark & --         \\
    \texttt{audio-only}    & \checkmark & \checkmark & \checkmark & --         & \checkmark \\
    \texttt{audio+acoustic}    & \checkmark & \checkmark & \checkmark & \checkmark & \checkmark \\
    \bottomrule
    \end{tabular}
\end{table}

\subsection{Models and prompting protocol}
\label{sec:models}

We evaluate three multimodal LLMs: \texttt{google/gemini-3.0-flash}, \texttt{openai/gpt-4o} for text conditions paired with \texttt{openai/gpt-4o-audio} for audio, and \texttt{qwen3.5 omni-plus}. The three models are tested on five conditions, which use the same structured-output schema. For each utterance, the model returns a JSON object containing a \texttt{ratings} field (one categorical label per dimension) and an \texttt{evidence} field (a scalar value or list of phones/stressed vowels per dimension, with a one-sentence reason). The four dimensions are presented in a per-utterance random order seeded by a deterministic hash, removing position-of-listing as a confound. We use API temperature $0$ and retry malformed responses up to three times. The full system and user prompts are in Appendix~\ref{app:prompts}.

\subsection{Evaluation metrics}
\label{sec:metrics}

We report three metrics per (model $\times$ condition $\times$ dimension) cell, designed to separate three questions: did the model assign the correct label, is its explanation internally coherent, and is the explanation valid against gold evidence?

\paragraph{Rating Accuracy (RA).}
RA compares the model's predicted rating against the gold label. For three-class dimensions we report macro-F1,
\begin{equation}
\text{RA}_{\text{cat3}} = \frac{1}{3}\sum_{c \in \mathcal{C}} \frac{2 P_c R_c}{P_c + R_c},
\end{equation}
with $\mathcal{C} = \{\text{slow}, \text{normal}, \text{fast}\}$ or $\{\text{monotone}, \text{normal}, \text{varied}\}$. For binary dimensions we report positive-class F1, $\text{RA}_{\text{bin}} = 2 P_1 R_1 / (P_1 + R_1)$, the standard choice for imbalanced detection tasks. We do not macro-average across dimensions, because they differ in label structure and base rate.

\paragraph{Evidence Coherence (EC).}
EC measures whether the model's explanation is internally coherent with the evidence it cites. The judge sees the rating, the cited evidence, and the reason, but \emph{not} the gold label. EC is scored on three bins, $\{0, 0.5, 1.0\}$, by an LLM judge with the prompt template in Appendix~\ref{app:prompts}. EC is high when the reason follows from the cited evidence on its own terms, even if the rating happens to be wrong against ground truth.

\paragraph{Grounded Correctness (GC).}
GC measures whether the explanation remains valid when gold evidence is considered. The LLM judge sees everything in EC plus the gold label and the dimension-specific gold evidence (duration and speaking-rate cues for fluency, F0-based cues for pitch variation, errored stressed vowels for stress, errored phones for phoneme). To keep GC parallel to EC, we report it on the same three-point scale, $\{0, 0.5, 1.0\}$: $1.0$ denotes a correct rating with valid evidence-based reasoning, $0.5$ a correct rating with decorative, unsupported or stereotype-driven reasoning, and $0.0$ an incorrect, unsupported response. This shared scale makes the EC/GC cross-tab in §\ref{sec:results_decoupling} directly interpretable and separate from the confabulation threshold defined below.

We use \texttt{google/gemini-2.5-pro} as the judge; RA is computed deterministically while the judge scores EC and GC only. The structured evidence packages and operationally defined three-bin rubric mitigate the subjectivity typical of LLM-as-judge settings~\citep{ye2024llmjudge_bias}.
  
  Because RA is a macro-averaged F1 score while EC and GC are means over an ordinal three-point scale ($\{0, 0.5, 1.0\}$), their absolute values are not
  on a common scale and should not be directly compared or subtracted. The three metrics are nonetheless complementary: high RA with low GC reveals that correct labels are assigned without grounded reasoning, and high EC with low GC is the signature of confabulation.


\paragraph{Confabulation rate.}
We additionally report a derived rate that combines EC and GC. We label a cell as having a coherent reason when $\mathrm{EC} \geq 0.7$ and a grounded rating when $\mathrm{GC} \geq 0.7$ (both correspond to the top bin on each judge scale). The confabulation rate is the share of cells with a coherent reason but an ungrounded rating:
\begin{equation}
\text{Conf} = \frac{\big|\{c : \mathrm{EC}_c \geq 0.7 \,\wedge\, \mathrm{GC}_c < 0.7\}\big|}{|\mathcal{C}|}.
\end{equation}
This identifies cells that look right under reference-match scoring but fail under ground-truth-aware scoring, the failure mode that motivated~\citet{turpin2023unfaithful} in the chain-of-thought setting.

\section{Results}
\label{sec:results}

\subsection{Rating accuracy and grounded reasoning decouple}
\label{sec:results_decoupling}

\begin{figure}[t]
\hspace*{-3mm}
  \includegraphics[width=1.05\columnwidth]{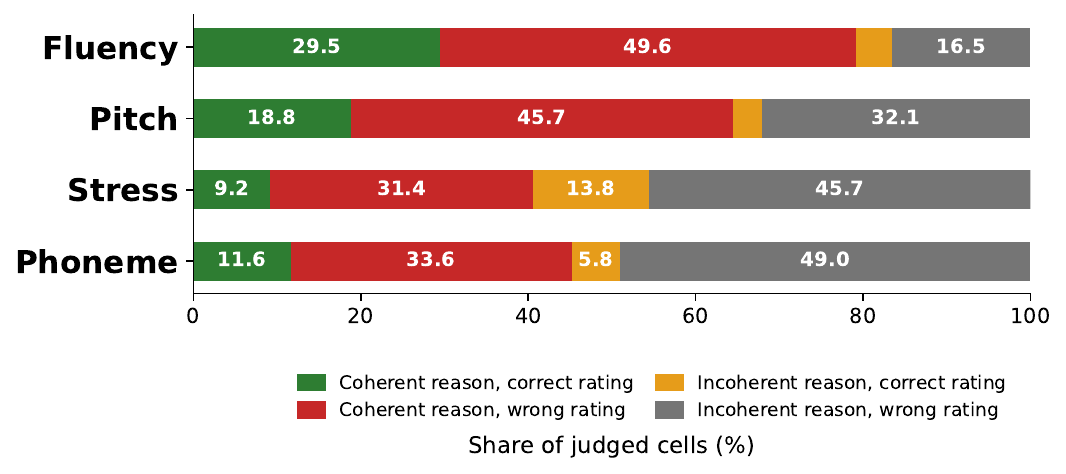}
  \caption{Distribution of reasoning coherence vs. rating correctness per pronunciation dimension. Across all dimensions, wrong yet coherent diagnoses (\textit{confabulation}, red bars) substantially outnumber genuinely grounded feedback (green bars).}
  \label{fig:confabulation}
\end{figure}

Across $34{,}887$ cells judged by both EC and GC, $55.4\%$ have a coherent reason ($\mathrm{EC} \geq 0.7$), but only $15.8\%$ pair a coherent reason with a grounded rating ($\mathrm{GC} \geq 0.7$). 
The remaining $39.6\%$ are coherent but wrong, corresponding to the confabulation cells defined in §\ref{sec:metrics}. 
Figure~\ref{fig:confabulation} shows the four-quadrant breakdown by pronunciation dimension. Confabulation consistently exceeds genuine grounding, with ratios of $1.7$ for fluency, $2.4$ for pitch variation, $2.9$ for phoneme correctness, and $3.4$ for stress correctness. The largest gap occurs for stress, where target alignment is hardest. This pattern echoes~\citet{turpin2023unfaithful}'s findings on unfaithful chain-of-thought reasoning, but is larger in magnitude: it persists even when a structured-output schema requires the model to provide typed evidence before generating its explanation.

This decoupling has a direct consequence for CAPT evaluation. A high RA score shows that the model selected the correct label but does not show that the explanation identifies the right acoustic or phonetic evidence. High EC shows that the explanation is well-formed relative to the prompt but does not show the rating is correct. GC is the relevant metric for feedback quality because it asks whether rating and explanation jointly support a valid learner-facing diagnosis. The remaining subsections decompose this overall pattern by phoneme inventory (§\ref{sec:results_phoneme_stereotype}), marginal rating distribution (§\ref{sec:results_default_bias}), and evidence condition (§\ref{sec:results_encoding_ladder}).

\subsection{Phoneme feedback collapses onto an L1-independent stereotype}
\label{sec:results_phoneme_stereotype}

If phoneme feedback were grounded in the learner's actual production, the LLM cited-phone distribution should vary with both the speaker's L1 and the prompting condition; this is the substitution pattern predicted by classical contrastive analysis~\cite{lado1957linguistics, eckman1977markedness, swan2001learner}. Instead, it collapses onto a small inventory of familiar L2-English difficulty phones. Table~\ref{tab:phoneme_errors} reports the pooled top-five over-claimed phones per L1, across models and evidence conditions ($15$ model-condition cells per L1).

\begin{table}[H]
\centering
\scriptsize
\setlength{\tabcolsep}{0pt}
\caption{Pooled top-five over-claimed phones per L1}

\renewcommand{\arraystretch}{1}
\begin{tabular}{l cccccc}
\toprule
Rank & Arabic & Hindi & Korean & Mandarin & Spanish & Vietnamese \\
\midrule
\textbf{\#1}
& \textbf{/\textipa{T}/} & \textbf{/\textipa{D}/} & \textbf{/\textipa{\*r}/} & \textbf{/\textipa{D}/} & \textbf{/\textipa{D}/} & \textbf{/\textipa{D}/} \\
& \tiny 14.5\% (+13.6) & \tiny 19.2\% (+12.4) & \tiny 15.9\% (+8.0) & \tiny 14.2\% (+3.1) & \tiny 15.2\% (+6.6) & \tiny 14.2\% (+5.5) \\
& \tiny 14/15 & \tiny 13/15 & \tiny 14/15 & \tiny 14/15 & \tiny 14/15 & \tiny 14/15 \\
\midrule
\textbf{\#2}
& \textbf{/\textipa{D}/} & \textbf{/\textipa{T}/} & \textbf{/\textipa{D}/} & \textbf{/\textipa{T}/} & \textbf{/\textipa{T}/} & \textbf{/\textipa{T}/} \\
& \tiny 18.3\% (+6.6) & \tiny 12.9\% (+12.0) & \tiny 7.8\% (+6.8) & \tiny 17.3\% (+14.3) & \tiny 15.4\% (+11.9) & \tiny 12.6\% (+11.6) \\
& \tiny 12/15 & \tiny 10/15 & \tiny 11/15 & \tiny 13/15 & \tiny 12/15 & \tiny 13/15 \\
\midrule
\textbf{\#3}
& \textbf{/\textipa{\*r}/} & \textbf{/\textipa{v}/} & \textbf{/\textipa{v}/} & \textbf{/\textipa{\*r}/} & \textbf{/\textipa{\*r}/} & \textbf{/\textipa{\*r}/} \\
& \tiny 9.7\% (+6.0) & \tiny 19.2\% (+19.2) & \tiny 9.6\% (+6.7) & \tiny 13.8\% (+9.3) & \tiny 9.2\% (+7.0) & \tiny 10.0\% (+2.0) \\
& \tiny 12/15 & \tiny 10/15 & \tiny 11/15 & \tiny 12/15 & \tiny 9/15 & \tiny 13/15 \\
\midrule
\textbf{\#4}
& \textbf{/\textipa{v}/} & \textbf{/\textipa{\*r}/} & \textbf{/\textipa{T}/} & \textbf{/\textipa{v}/} & \textbf{/\textipa{v}/} & \textbf{/\textipa{z}/} \\
& \tiny 13.3\% (+5.6) & \tiny 14.5\% (+3.8) & \tiny 15.5\% (+13.4) & \tiny 12.6\% (+8.6) & \tiny 10.6\% (+6.8) & \tiny 11.0\% (+5.4) \\
& \tiny 11/15 & \tiny 10/15 & \tiny 9/15 & \tiny 9/15 & \tiny 9/15 & \tiny 6/15 \\
\midrule
\textbf{\#5}
& /\textipa{p}/ & /\textipa{w}/ & /\textipa{f}/ & /\textipa{d}/ & /\textipa{s}/ & /\textipa{I}/ \\
& \tiny 23.6\% (+20.4) & \tiny 11.9\% (+9.0) & \tiny 14.1\% (+14.1) & \tiny 5.6\% (+1.6) & \tiny 4.8\% (+3.9) & \tiny 5.6\% (+0.2) \\
& \tiny 5/15 & \tiny 5/15 & \tiny 4/15 & \tiny 6/15 & \tiny 5/15 & \tiny 5/15 \\
\bottomrule
\end{tabular}
\begin{minipage}{\linewidth}
\footnotesize
\emph{Note.} Entries show phone, claim rate, mean model--gold gap in parentheses, and recurrence among the top-five over-claimed phones across 15 cells per L1. Phones are ranked by recurrence, then by mean gap.
\end{minipage}
\label{tab:phoneme_errors}
\end{table}

The same four phones, /\textipa{T}/, /\textipa{D}/, /\textipa{\*r}/, and /\textipa{v}/, dominate the top-five over-claim list for every L1, despite L1-specific gold-error distributions. /\textipa{D}/ appears in $14/15$ cells for Mandarin and Spanish; /\textipa{T}/ appears in $14/15$ cells for Arabic; the two dental fricatives (/\textipa{T}/, /\textipa{D}/) plus /\textipa{\*r}/ occupy three of the top four ranks for Mandarin, Spanish, and Vietnamese. The pattern is not driven by a single model or prompting configuration, because the recurrence counts pool across all $15$ model-condition cells. Aggregated across cells, /\textipa{T}/, /\textipa{D}/, and /\textipa{\*r}/ jointly account for over a third of all emitted phoneme tokens, against approximately $20\%$ of the gold-error distribution. Richer evidence does not eliminate the effect: IPA, acoustic, and audio-based conditions all produce the same inventory, with the condition-level matrix in Appendix Table ~\ref{tab:appendix_phoneme}.

This contradicts both the L1-specific substitution predictions of contrastive analysis~\cite{eckman1977markedness, swan2001learner} and the speaker-adaptive behaviour reported by recent multimodal-LLM graders~\cite{fu2024multimodalpa, ma2025l2speechllm}. A learner who receives this feedback is told a particular phone was mispronounced, but the cited phone often reflects a generic L2-English-difficulty prior rather than a phone the speaker actually missed.

\subsection{Prosodic predictions default to L2-stereotype classes}
\label{sec:results_default_bias}

A marginal-distribution check on the rating output reveals a pattern not visible from RA alone. On every prosodic dimension at \texttt{text-only}, all three models assign one class to the large majority of utterances, and the over-emitted class is the same across models (Table~\ref{tab:default_bias}). The over-emitted classes are \texttt{slow} for fluency, \texttt{monotone} for pitch variation, and \texttt{1\,=\,error} for stress correctness; each one matches a popular L2-English stereotype, the kind of fixed parametric prior that~\citet{mallen2023llms} and~\citet{tao2024contextparametric} flag as the default fallback when supplied context is weak.

\begin{table}[t]
\centering
\caption{Default-class prediction rates for prosodic dimensions by model and evidence condition.}

\scriptsize
\setlength{\tabcolsep}{1.5pt}
\renewcommand{\arraystretch}{1.5}
\begin{tabular}{ll ccc}
\hline
\textbf{Dimension} & \textbf{Condition} & \textbf{Gemini 3.0} & \textbf{GPT-4o} & \textbf{Qwen 3.5 Omni} \\
(\texttt{class}, \textcolor{blue}
{gold rate}) & & & & \\
\hline
\textbf{Fluency} & \texttt{text-only}       & 72\% & 70\% & 84\% \\
(\texttt{slow}, \textcolor{blue}{33\%}) & \texttt{text+ipa}            & 74\% & 66\% & 91\% \\
                      & \texttt{text+acoustic}             & 49\% & 64\% & 22\% \\
                      & \texttt{audio-only}          & 19\% &  8\% & 49\% \\
                      & \texttt{audio+acoustic}      & \textbf{32\%} & \textbf{32\%} & \textbf{34\%} \\
\hline
\textbf{Pitch}   & \texttt{text-only}       & 66\% & 62\% & 95\% \\
(\texttt{monotone}, \textcolor{blue}{36\%}) & \texttt{text+ipa}            & 65\% & 63\% & 97\% \\
                      & \texttt{text+acoustic}             & \textbf{1\%} &  \textbf{0\%} &  \textbf{0\%} \\
                      & \texttt{audio-only}          & 34\% &  2\% & 86\% \\
                      & \texttt{audio+acoustic}      &  1\% &  0\% &  0\% \\
\hline
\textbf{Stress}  & \texttt{text-only}       & 82\% & 95\% & 96\% \\
(\texttt{1=error}, \textcolor{blue}{4\%})  & \texttt{text+ipa}            & 85\% & 95\% & 98\% \\
                      & \texttt{text+acoustic}             & \textbf{52\%} & \textbf{90\%} & \textbf{56\%} \\
                      & \texttt{audio-only}          & 41\% & 82\% & 66\% \\
                      & \texttt{audio+acoustic}      & 45\% & 82\% & 35\% \\
\hline
\end{tabular}
\begin{minipage}{0.95\linewidth}
\footnotesize
\emph{Note.} For each dimension, the class in parentheses is the default class whose prediction rate is reported; the gold rate gives its prevalence in the reference labels.
\end{minipage}

\vspace{0.5em}
\label{tab:default_bias}
\vspace{-0.2in}
\end{table}

At \texttt{text-only}, all three models over-predict the stereotype class: \texttt{slow} on $70$--$84\%$ of utterances against a gold rate of $33\%$; \texttt{monotone} on $62$--$95\%$ against $36\%$; \texttt{1\,=\,error} on $82$--$96\%$ against $4\%$. The bias is uniform in direction even though the gold distribution points the opposite way on fluency (most utterances are \texttt{normal} or \texttt{fast}) and pitch variation (most are \texttt{varied}).

Conditions diverge in their effect on this concentration. IPA leaves the over-emission essentially unchanged on every dimension. On pitch variation, \texttt{text+acoustic} collapses the \texttt{monotone} rate to $0$--$1\%$ for all three models, an unambiguous abandonment of the stereotype class; \texttt{audio+acoustic} produces the same effect. On fluency, \texttt{text+acoustic} reduces the \texttt{slow} rate substantially for Gemini ($72 \to 49$) and dramatically for Qwen ($84 \to 22$), but barely for GPT-4o ($70 \to 64$). On stress, the error rate reduces modestly under \texttt{text+acoustic} and \texttt{audio+acoustic} but never approaches the $4\%$ gold rate. Per-utterance grounding therefore tracks the specificity of the supplied evidence to the target feature, not the richness of the evidence package, a sharper version of the context-versus-parametric result of~\citet{tao2024contextparametric}.

\subsection{Acoustic evidence helps pitch and fluency, not stress and phoneme}
\label{sec:results_encoding_ladder}

\begin{figure*}[t]
    \centering
    \includegraphics[
        width=\textwidth,
        height=0.4\textheight,
        keepaspectratio
    ]{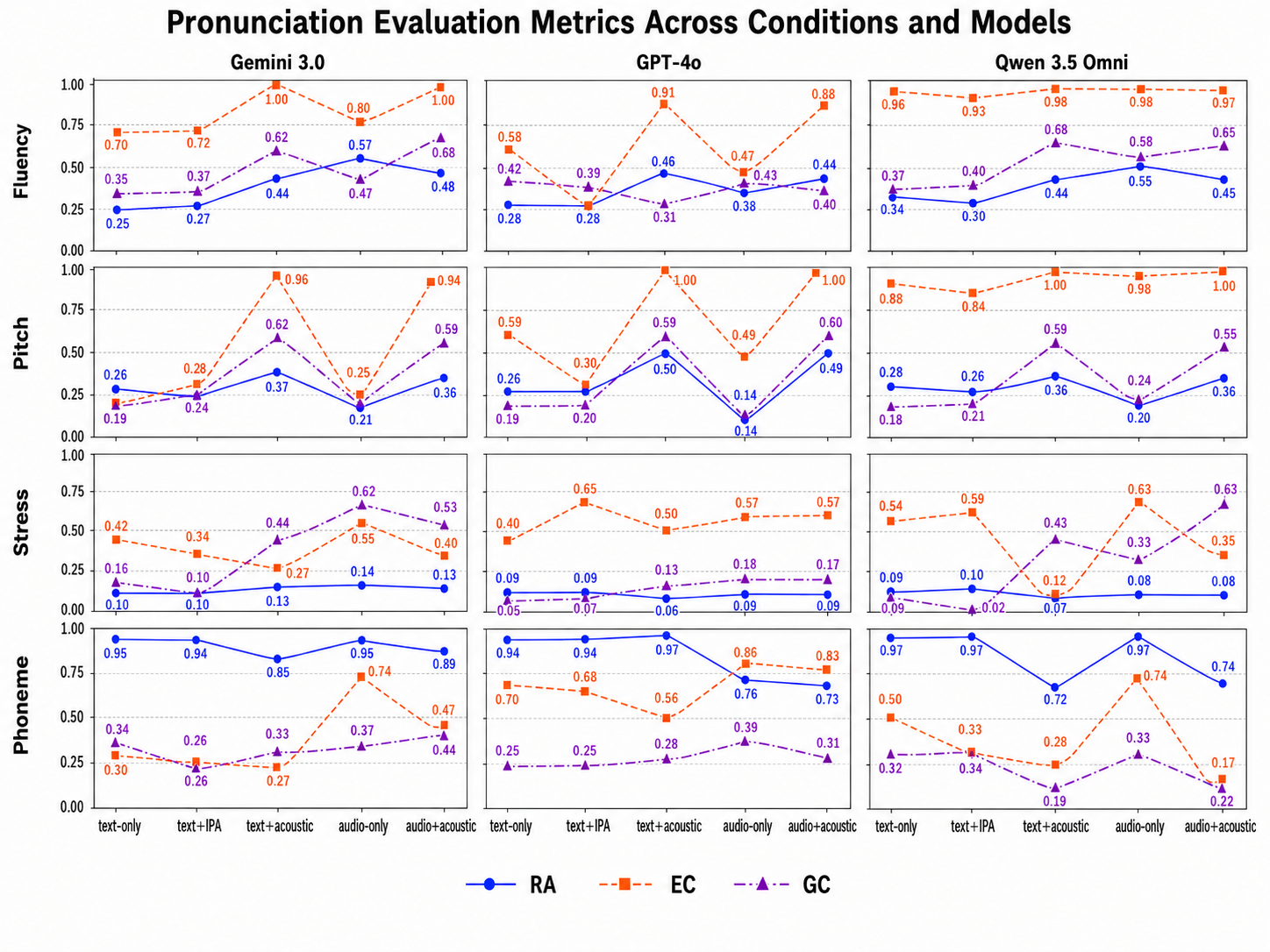}

    \vspace{-1em}
    \captionsetup{justification=justified}
    \caption{Pronunciation evaluation metrics across models and prompt conditions. Each panel tracks RA, EC, and GC for a specific model-dimension pair, with selected exact values annotated. RA denotes Rating $F_1$, while EC and GC represent Evidence Coherence and Grounded Correctness, respectively.}
    \label{fig:metrics}
\end{figure*}

Figure~\ref{fig:metrics} reports RA, EC, and GC per 
(model $\times$ condition $\times$ dimension). The pattern is direct: acoustic evidence improves the rating only when the supplied feature directly probes the target. F0 range directly probes pitch variation; duration plus word count probes fluency through a single arithmetic step; no supplied measurement directly probes which phone or which syllable was wrong, so stress and phoneme correctness remain ungrounded.

\paragraph{Pitch variation.} Pitch variation provides a clear case where explicit acoustic evidence improves grounding. GC rises from ($0.18$--$0.19$) to ($0.45$--$0.62$) across the three models when textualised F0 values are supplied (\texttt{text+acoustic}). The \texttt{audio+acoustic} condition shows a similar gain ($0.55$--$0.61$), whereas \texttt{audio-only} remains weak, indicating that the effective intervention is the explicit textual representation of F0, not audio by itself. The pattern matches the gap identified in speech-LM roadmaps between perceiving non-semantic acoustic cues and using them as evidence for diagnostic reasoning~\cite{bu2024sagi,cui2024speechlm_survey}.
\paragraph{Fluency.}
Fluency shows a similar but less uniform pattern. Under \texttt{text+acoustic}, Gemini and Qwen lift in GC from $0.27$ to $0.47$ and from $0.25$ to $0.54$. GPT-4o does not (drops from $0.35$ to $0.27$). The supplied measurement is utterance duration, which requires the model to compute words per second from the target sentence. The split-by-model pattern is consistent with the finding that small arithmetic derivations are not uniformly solved across models~\cite{turpin2023unfaithful}, and it shows that even feature-readout tasks are not solved when the readout step is non-trivial.

\paragraph{Stress correctness.}
Stress correctness presents the hardest grounding problem. RA stays low across all conditions ($0.06$--$0.14$) and GC never exceeds $0.61$. The ceiling is architectural: grounding a stress judgment requires the model to know which syllable should be stressed, locate the stress produced by the speaker, and compare the two — a target-to-realisation alignment that acoustic summaries alone cannot supply.

\paragraph{Phoneme correctness.}
As established in §\ref{sec:results_phoneme_stereotype}, phoneme RA is near-ceiling ($\geq 0.94$) due to the $96\%$ base rate, while GC remains low ($0.19$--$0.38$) because cited phones are L1- and condition-independent. No evidence condition closes this gap: the model can detect that \emph{some} error exists but cannot ground \emph{which} phone was mispronounced. 

  \label{sec:results_gender}

\subsection{Metadata Ablation Study}
 In the previous experiments, every prompt includes the speaker's L1 and gender. To test whether
  models use these labels rather than the speech itself, we ran a
  \texttt{metadata\_suppression} variant: the same evidence, plus one
  instruction: ``You MUST NOT use the speaker's L1 background or
  gender as a basis for any judgement.'' Because both variants share
  identical evidence, any score change is caused by the label alone.
  
   The effect is dimension-specific. Stress correctness shows the clearest effect. With the L1 label visible, three of four models score only 0.08--0.14 on stress, essentially flagging a stress error on almost every utterance,
  whether or not one actually occurred. Once the label is hidden, scores jump to 0.40--0.65, with Grounded Correctness rising in parallel. Obviously, the model was not listening to the speech; it was seeing
  ``L2 learner'' and assuming errors.
  
  Pitch variation reveals a gender effect. With gender visible, female speakers score 0.16--0.24 higher than male speakers; models routinely label male speakers as \texttt{monotone} even when the
  gold annotations show the opposite. Hide the gender label and the gap reverses: male speakers now score 0.13--0.27 higher than female.
  The model was responding to the gender tag, not to the speaker's actual pitch.

  Phoneme correctness and fluency do not change when labels are removed. For phoneme correctness this is expected: $96\%$ of utterances actually contain a phoneme error, so models score near-ceiling (0.94--0.97) regardless of what the label says, and
  the over-claimed phone inventory
  (§\ref{sec:results_phoneme_stereotype}) is L1-independent anyway.
  For fluency, there is no consistent stereotype linking L1 or gender
  to speech rate, so removing either label changes nothing.

Together, these results answer RQ3. Where a widely held stereotype
  exists, including stress errors for L2 learners, monotone pitch for male
  speakers, the demographic label overrides the acoustic evidence.
  Where no such stereotype exists, as with phoneme inventory and
  speech rate, it has no effect.

\section{Discussion}
\label{sec:disc}

The empirical findings reveal a fundamental limitation in how audio-capable LLMs process speech: a severe dissociation between surface-level task performance (Rating Accuracy) and actual factual grounding (Grounded Correctness). This gap is not random noise but a structural vulnerability. When explicit acoustic evidence is absent or difficult to parse, LLMs do not fail gracefully by expressing uncertainty; instead, they default to demographic and pedagogical priors ingrained during pretraining.

This failure mode is dimension-specific, governed by the distinction between \emph{feature readout} and \emph{target-to-realisation alignment}. For fluency and pitch variation, low-level acoustic summaries map almost directly onto the required label. Diagnosing phoneme errors and lexical stress deviations, by contrast, requires comparing the target canonical form with the learner’s actual realisation, determining not just what the signal sounds like but where it deviates from a language-specific expectation. Acoustic summaries cannot encode this comparative alignment; in their absence, the model fills the gap with prior-driven confabulation. This finding is consistent with recent work showing that prosodic sensitivity in speech-language models remains cue-dependent rather than uniformly reliable~\cite{qian2025prosodylm, chen2025textpa}.

The errors follows a predictable pattern. Phoneme explanations collapse onto the rigid pretraining inventory documented in §\ref{sec:results_phoneme_stereotype}, irrespective of the speaker’s L1 or evidence richness, while demographic labels selectively override evidence on the two dimensions where a plausible stereotype exists. These priors superficially resemble valid cross-linguistic transfer knowledge but become harmful when applied to individual learners without evidence of their actual errors, the failure mode identified by~\citet{turpin2023unfaithful} and the parametric-knowledge literature~\cite{petroni2019language, kassner2021negated, mallen2023llms}.

Two implications follow for CAPT systems. First, evaluation standards must evolve. High $F_1$ is an illusion of competence, and future benchmarks must decouple task accuracy from reasoning validity by reporting multi-tiered grounding metrics such as EC and GC. Second, architectures must be refactored. A more robust design uses specialised acoustic models for target-to-realisation alignment and deploys LLMs strictly as downstream natural-language interfaces.

\section{Conclusion}
This study tested whether audio-capable LLMs can ground L2 pronunciation feedback in objective phonetic evidence. Our analysis quantifies the decoupling: $39.6\%$ of evaluated outputs are structurally coherent but entirely ungrounded, while only $15.8\%$ achieve both internal coherence and factual grounding. Phoneme feedback collapses onto a
fixed L2-difficulty inventory irrespective of the
speaker’s L1 background or the evidence richness. Prosodic predictions show a similar prior-driven pattern through single-class clustering and demographic-prior effects. These systematic, prior-driven biases demonstrate that current LLMs remain unreliable as autonomous pronunciation diagnostic engines. They are far more dependable when confined to verbalising externally verified, structurally aligned speech evidence.

\section{Limitations}
\label{sec:limitations}

The evaluation is limited to three general-purpose multimodal systems and whether these patterns generalize to specialized speech-language models or spontaneous L2 speech remains an open question. Sample sizes for certain per-L1 groups and binary-error categories in the L2-ARCTIC corpus are relatively small. The fluency and pitch labels are derived from human-verified TextGrids rather than independent human annotation. The LLM-as-a-judge approach introduces heuristic biases that may affect absolute scores; however, these factors do not readily explain our primary findings: phoneme feedback repeatedly relies on generic L2-English error stereotypes, and phoneme and stress feedback remain weakly grounded even when acoustic evidence is supplied. 

\bibliography{custom_claude2}

\appendix
\twocolumn

\section{System and User Prompt Templates}
  \label{app:prompts}

  All experiments use one of three system messages combined with a per-condition user message. The three system messages share a common output-format clause and differ only in their framing of the task.

  \subsection*{System messages}

\paragraph{\textsc{system\_default}} 

\begin{quote}\small
  You are a phonetician evaluating L2 English pronunciation. For each of the four dimensions listed below, pick the categorical label that best characterises the speaker's production relative to native English
  norms. \textit{[shared output rules]}
  \end{quote}

\paragraph{\textsc{system\_grounded}} for \texttt{text+acoustic} and \texttt{audio+acoustic}.

  \begin{quote}\small
  You are a phonetician evaluating L2 English pronunciation. For each of the four dimensions, pick the categorical label that best characterises the speaker's production relative to native English norms. Base your
  decisions on the acoustic measurements and/or the audio provided in the EVIDENCE block. In the \texttt{value\_used} / \texttt{vowels\_used} / \texttt{phonemes\_used} fields, report the actual measurement or
  symbols you consulted (do not invent default values). \textit{[shared output rules]}
  \end{quote}

  \paragraph{Shared output rules.}

  \begin{quote}\small
  For every dimension, fill BOTH: \texttt{rating} (the categorical label from the allowed set) and \texttt{evidence} (a typed value or list of phoneme symbols, plus a one-sentence \texttt{reason}). Even if you are
  uncertain, you MUST commit to a rating; do not leave any field empty. Return ONLY a single JSON object that matches the schema, with no prose and no markdown fences.
  \end{quote}

The four dimensions are listed in a per-utterance randomised order, seeded reproducibly by
  \texttt{hashlib(utt\_id, condition)}.

  \begin{quote}\small\ttfamily
  Dimensions to rate (in any order):\\
  \hspace*{1em}- Fluency / speaking rate (label: "slow" | "normal" | "fast")\\
  \hspace*{1em}- Pitch variation / F0 range (label: "monotone" | "normal" | "varied")\\
  \hspace*{1em}- Stress correctness (0 = no stress errors, 1 = at least one stress error)\\
  \hspace*{1em}- Phoneme correctness (0 = no phoneme errors, 1 = at least one phoneme error)\\

  \rmfamily --- EVIDENCE ---\\
  \ttfamily Speaker L1 background: Mandarin\\
  Speaker gender: Female\\
  Target sentence: "The quick brown fox jumps over the lazy dog."\\

  Canonical phoneme alignment (one phoneme per line, start\_sec--end\_sec : PHONEME):\\
  0.00--0.10 : DH\\
  0.10--0.18 : AH\\
  0.18--0.32 : K\\
  \ldots\\
  \hspace*{2em}- Duration:\hspace*{2em}2.84 s\hspace*{1em}(tip: divide words by duration for words/sec)\\
  \hspace*{2em}- F0 min:\hspace*{2em} 88.3 Hz\\
  \hspace*{2em}- F0 max:\hspace*{2em}243.1 Hz\hspace*{1em}(tip: F0 range = max -- min)\\
  \hspace*{2em}- Intensity range: 14.6 dB\\
  \rmfamily --- END EVIDENCE ---\\

  \ttfamily JSON schema you must return: \{ "ratings": \{\ldots\}, "evidence": \{\ldots\} \}
  \end{quote}

\paragraph{\textsc{system\_judge}}
  \begin{quote}\small
    "You audit a phonetician's reasoning against ground truth. For a single "
    "pronunciation dimension you receive: (a) the task and prompt the phonetician "
    "was shown, (b) the rating they assigned, (c) the evidence they cited, "
    "(d) the one-sentence reason they wrote, and (e) the ground-truth label "
    "plus the underlying GT value or error list. Your job is to return two "
    "scores: a binary rating-correct flag and a 3 level grounded-correctness "
    "(GC) score. The GC rubric distinguishes (i) correct rating with sound "
    "evidence-based reasoning, (ii) correct rating with decorative or "
    "stereotype-driven reasoning (lucky guess), (iii) wrong rating with no defensible reading. "
    "Return ONLY the JSON object."
\end{quote}
\begin{quote}\small\ttfamily
JUDGE\_USER\_TEMPLATE = """DIMENSION: \{dim\_label\}\\
EVIDENCE SHOWN TO PHONETICIAN:\\
\{evidence\_block\}\\
PHONETICIAN'S OUTPUT FOR THIS DIMENSION:\\
\hspace*{2em}RATING:\hspace*{6.5em} \{rating\}\\
\hspace*{2em}EVIDENCE CITED (\{field\}): \{cited\}\\
\hspace*{2em}REASON:\hspace*{5.5em} \{reason\}\\
\\
GROUND TRUTH FOR THIS UTTERANCE:\\
\hspace*{2em}GT LABEL:\hspace*{5.5em} \{gt\_label\}\\
\hspace*{2em}GT \{gt\_aux\_name\}:\hspace*{4.5em} \{gt\_aux\}\\
\\
Scoring:\\
\hspace*{2em}rating\_correct: 1 if RATING equals GT LABEL (string-equal for categorical; integer-equal for binary); 0 otherwise.\\
\\
\hspace*{2em}gc — Grounded Correctness (3-point):\\
\hspace*{2em}1.0: Rating matches GT AND reason is a valid explanation for the correct rating, using the cited evidence.\\
\hspace*{2em}0.5: Rating matches GT but reason is decorative, wrong, or invokes facts not in evidence (correct label, ungrounded rationale).\\
\hspace*{2em}0.0: Rating disagrees with GT and the reason does not support any defensible reading.\\
Return ONLY JSON: \{\{"rating\_correct": 0|1, "gc": <1.0|0.5|0.0>, "verdict": "<one sentence>"\}\}\\
"""
\end{quote}

\section {Ranked top-five phones claimed as mispronounced across  models  $\times$ condition}

\begin{table*}[t]

\centering
\caption{Ranked top-five phones claimed as mispronounced across three models, five prompting conditions, and six L1 backgrounds. Each cell reports items ordered down from Rank 1 to 5. Format follows: Phone (Percentage\,\% / Model--Gold Gap $\Delta$). Empty spaces tracking unavailable data indicators for non-speech setups are formatted as (--).}
\scriptsize
\setlength{\tabcolsep}{3.5pt}
\renewcommand{\arraystretch}{0.6}
\begin{tabular}{ll p{2.1cm}p{2.1cm}p{2.1cm}p{2.1cm}p{2.1cm}p{2.1cm}}
\toprule
Model & Condition & Arabic & Hindi & Korean & Mandarin & Spanish & Vietnamese \\
& & \multicolumn{6}{c}{\textit{Ranked Top-5 claimed phones formatted as: Phone (Percentage\,\% / $\Delta$)}} \\
\midrule
\textbf{Gemini} & \texttt{baseline} & \textbf{P},(31.5/+28.1) \par \textbf{V},(26.6/+18.9) \par \textbf{DH},(8.7/-2.3) \par \textbf{TH},(7.1/+6.2) \par \textbf{R},(2.9/-1.3) & \textbf{V},(30.8/+30.8) \par \textbf{DH},(18.6/+11.2) \par \textbf{W},(16.6/+13.4) \par \textbf{T},(5.3/-6.4) \par \textbf{TH},(4.9/+3.8) & \textbf{Z},(21.5/+4.1) \par \textbf{F},(18.2/+18.2) \par \textbf{R},(12.4/+4.8) \par \textbf{L},(10.2/+9.1) \par \textbf{V},(9.5/+6.2) & \textbf{V},(20.7/+17) \par \textbf{TH},(18.4/+14.7) \par \textbf{DH},(16.2/+4.1) \par \textbf{L},(12.8/+5.4) \par \textbf{R},(9/+3.9) & \textbf{V},(17.7/+13.5) \par \textbf{Z},(13.8/+0.1) \par \textbf{IH},(11/-0.9) \par \textbf{DH},(9.8/+0.9) \par \textbf{TH},(8.3/+4.7) & \textbf{DH},(14.8/+5.9) \par \textbf{Z},(12.7/+7.3) \par \textbf{TH},(11/+9.6) \par \textbf{L},(10.6/+3.4) \par \textbf{D},(8.5/-0.5) \\
& \texttt{ipa} & \textbf{P},(29.8/+26) \par \textbf{V},(20.2/+12.6) \par \textbf{DH},(11.5/+0.1) \par \textbf{TH},(6.7/+5.8) \par \textbf{IH},(3.6/+0.7) & \textbf{V},(31.5/+31.5) \par \textbf{DH},(15.7/+8.3) \par \textbf{W},(13.2/+10) \par \textbf{TH},(9.8/+8.7) \par \textbf{T},(7.7/-4) & \textbf{Z},(19.9/+2.5) \par \textbf{R},(14.7/+7.1) \par \textbf{F},(13.6/+13.6) \par \textbf{V},(9.9/+6.7) \par \textbf{L},(9.2/+8.1) & \textbf{TH},(19.9/+15.8) \par \textbf{V},(19.5/+15.9) \par \textbf{DH},(15.4/+3.7) \par \textbf{R},(13.5/+8.4) \par \textbf{L},(11.2/+4.2) & \textbf{DH},(15.2/+6.2) \par \textbf{V},(12.9/+8.7) \par \textbf{Z},(10.2/-3.5) \par \textbf{IH},(9.1/-2.8) \par \textbf{TH},(7.6/+4) & \textbf{DH},(16.1/+7.1) \par \textbf{Z},(12.8/+7.4) \par \textbf{TH},(9.5/+8.1) \par \textbf{D},(9.5/+0.5) \par \textbf{V},(8/+5.3) \\
& \texttt{acoustic} & \textbf{P},(28.2/+26) \par \textbf{V},(17.2/+7.4) \par \textbf{DH},(8.6/-2.4) \par \textbf{TH},(5.2/+4.1) \par \textbf{IY},(4.6/+4.6) & \textbf{V},(27.1/+27.1) \par \textbf{W},(16.9/+14.7) \par \textbf{TH},(8.4/+7.3) \par \textbf{DH},(7.8/+1.2) \par \textbf{T},(6.6/-3.3) & \textbf{Z},(16.9/-1.5) \par \textbf{F},(13.4/+13.4) \par \textbf{R},(13.4/+3.4) \par \textbf{L},(10.5/+10.5) \par \textbf{V},(8.7/+5.4) & \textbf{TH},(14.5/+11.1) \par \textbf{V},(14/+9.9) \par \textbf{DH},(11.6/-0.6) \par \textbf{L},(7/+0.2) \par \textbf{D},(5.2/+1.2) & \textbf{V},(13.2/+9.8) \par \textbf{Z},(12.7/+0.4) \par \textbf{IH},(9/-3.4) \par \textbf{DH},(7.1/-1.8) \par \textbf{TH},(5.2/+0.4) & \textbf{TH},(9.5/+8.1) \par \textbf{Z},(8.3/+2.3) \par \textbf{DH},(7.9/-0.4) \par \textbf{R},(7.4/-0.8) \par \textbf{D},(7/-0.7) \\
& \texttt{audio-only} & \textbf{P},(15.7/+12.2) \par \textbf{DH},(11.8/+1.2) \par \textbf{V},(8.7/+0.8) \par \textbf{TH},(7.9/+7) \par \textbf{D},(4.4/-5.4) & \textbf{V},(32.2/+32.2) \par \textbf{DH},(14/+6.7) \par \textbf{W},(9.8/+6.7) \par \textbf{D},(8.9/+1.6) \par \textbf{T},(8.4/-3) & \textbf{Z},(11.4/-6.4) \par \textbf{V},(11.4/+8.5) \par \textbf{F},(11/+11) \par \textbf{R},(10.5/+2.6) \par \textbf{DH},(9/-10.8) & \textbf{DH},(15.8/+3.1) \par \textbf{TH},(11.5/+7.5) \par \textbf{V},(10.8/+6.7) \par \textbf{IH},(6.2/-1.9) \par \textbf{L},(6.2/-1.9) & \textbf{TH},(9.8/+6) \par \textbf{DH},(9.4/+0.1) \par \textbf{V},(9.4/+5.6) \par \textbf{Z},(9.4/-3.7) \par \textbf{IH},(9.1/-2.5) & \textbf{Z},(12.5/+6.6) \par \textbf{DH},(12.1/+3.3) \par \textbf{D},(10.6/+2.2) \par \textbf{TH},(7.9/+6.7) \par \textbf{L},(7.5/+0.8) \\
& \texttt{audio+acoustic} & \textbf{P},(12.9/+9.6) \par \textbf{DH},(10.8/0) \par \textbf{TH},(7.2/+6.1) \par \textbf{R},(5.7/+1.3) \par \textbf{V},(5.2/-3.5) & \textbf{V},(19/+19) \par \textbf{DH},(13/+5.5) \par \textbf{D},(8.2/-0.4) \par \textbf{T},(8.2/-2.6) \par \textbf{TH},(8.2/+7.1) & \textbf{R},(13.7/+4) \par \textbf{V},(8.3/+4.2) \par \textbf{Z},(8.3/-9.7) \par \textbf{L},(8.3/+8.3) \par \textbf{DH},(7.1/-10.9) & \textbf{D},(9/+4.8) \par \textbf{TH},(7.8/+4.8) \par \textbf{DH},(7.5/-5.2) \par \textbf{R},(6.3/+2.1) \par \textbf{L},(5.9/-2.6) & \textbf{IH},(8.6/-3.5) \par \textbf{Z},(7.8/-4.9) \par \textbf{DH},(7.8/-1.5) \par \textbf{TH},(7/+2.9) \par \textbf{V},(5.7/+2.3) & \textbf{DH},(8.4/0) \par \textbf{Z},(8/+2.7) \par \textbf{D},(7.6/-0.3) \par \textbf{P},(6.1/+3) \par \textbf{R},(5.7/-2.7) \\
\midrule
\textbf{GPT-4o} & \texttt{baseline} & \textbf{T},(6.9/+4.2) \par \textbf{S},(6.7/+6.7) \par \textbf{D},(6.1/-3) \par \textbf{R},(5.9/+2.2) \par \textbf{H},(5/+5) & \textbf{IH},(6/+1.6) \par \textbf{D},(5.8/-1.9) \par \textbf{R},(5.8/-4.1) \par \textbf{T},(5.3/-6.8) \par \textbf{AE},(4.9/+4.9) & \textbf{R},(7.2/-0.4) \par \textbf{L},(5.1/+4) \par \textbf{T},(4.9/-0.5) \par \textbf{D},(4.7/+0.4) \par \textbf{IH},(4.5/+1.3) & \textbf{R},(8/+2.8) \par \textbf{T},(5.2/-1) \par \textbf{IH},(5/-1.7) \par \textbf{H},(4.9/+4.9) \par \textbf{L},(4.6/-3.2) & \textbf{R},(6.4/+3.9) \par \textbf{D},(5.1/+0.7) \par \textbf{IH},(5.1/-7.3) \par \textbf{S},(4.5/+3.3) \par \textbf{T},(4.1/-0.8) & \textbf{T},(5.7/+2.2) \par \textbf{R},(5.3/-3.7) \par \textbf{D},(4.8/-4.1) \par \textbf{IH},(4.8/-1) \par \textbf{AH},(4.6/-1.2) \\
& \texttt{ipa} & \textbf{D},(5.7/-3.4) \par \textbf{S},(5/+5) \par \textbf{R},(4.6/+0.9) \par \textbf{T},(4.6/+1.9) \par \textbf{N},(4.6/+3.7) & \textbf{T},(5.9/-5.8) \par \textbf{R},(5.6/-5) \par \textbf{D},(5/-2.5) \par \textbf{AH},(4.7/+2.6) \par \textbf{N},(4.4/+1.2) & \textbf{T},(5.4/-0.2) \par \textbf{D},(5.2/+0.7) \par \textbf{R},(4.9/-2.8) \par \textbf{S},(4.5/+4.5) \par \textbf{L},(4.3/+3.2) & \textbf{R},(6.2/+1.4) \par \textbf{D},(4.9/+0.7) \par \textbf{T},(4.9/-0.4) \par \textbf{AH},(4.6/-1.2) \par \textbf{IH},(4.6/-2.8) & \textbf{D},(5.3/+0.4) \par \textbf{S},(5.3/+4.1) \par \textbf{T},(5.1/+0.2) \par \textbf{R},(5.1/+2.6) \par \textbf{N},(4.7/+1) & \textbf{D},(4.7/-4) \par \textbf{T},(4.5/+0.9) \par \textbf{R},(4.5/-4.6) \par \textbf{AH},(4.4/-1.5) \par \textbf{IH},(4.4/-1.1) \\
& \texttt{acoustic} & \textbf{AH},(7.8/+1.9) \par \textbf{IH},(5.9/+3.3) \par \textbf{T},(4.9/+2.4) \par \textbf{D},(4.4/-4.9) \par \textbf{R},(4.4/+0.2) & \textbf{AH},(7.6/+4.1) \par \textbf{IH},(7/+3.6) \par \textbf{R},(5.5/-5.8) \par \textbf{DH},(5.5/-1.5) \par \textbf{D},(4.4/-2.5) & \textbf{AH},(7.3/+0.5) \par \textbf{R},(5.4/-1.4) \par \textbf{IH},(5.2/+2.3) \par \textbf{DH},(5/-16.3) \par \textbf{T},(4.7/-1.2) & \textbf{AH},(6.9/+1.8) \par \textbf{D},(5.6/+1.7) \par \textbf{R},(5.5/+1.5) \par \textbf{DH},(5.1/-8) \par \textbf{IH},(5.1/-2.8) & \textbf{AH},(8.2/+0.9) \par \textbf{DH},(6.6/-3) \par \textbf{IH},(6.1/-5.7) \par \textbf{R},(5.4/+3.2) \par \textbf{ER},(4.3/+3.2) & \textbf{AH},(7.9/+1.6) \par \textbf{IH},(6.1/+0.6) \par \textbf{R},(4.9/-4) \par \textbf{DH},(4.7/-3.8) \par \textbf{D},(4/-4.1) \\
& \texttt{audio-only} & \textbf{T},(5.7/+2.9) \par \textbf{S},(5.5/+5.5) \par \textbf{R},(5.2/+1) \par \textbf{D},(5.2/-4.6) \par \textbf{IH},(4.7/+0.5) & \textbf{IH},(6.6/+4) \par \textbf{R},(4.7/-7) \par \textbf{D},(4.5/-2) \par \textbf{L},(4.3/-0.9) \par \textbf{T},(4.3/-4.8) & \textbf{IH},(6.6/+3.6) \par \textbf{R},(4.8/-2.5) \par \textbf{D},(4.7/+1.7) \par \textbf{T},(4.7/-1.2) \par \textbf{AH},(4.5/-2.9) & \textbf{IH},(7.3/-0.4) \par \textbf{AE},(4.9/+4.2) \par \textbf{S},(4.8/+4.8) \par \textbf{DH},(4.6/-7) \par \textbf{D},(4.5/-0.2) & \textbf{S},(6/+5.1) \par \textbf{IH},(5.8/-3.6) \par \textbf{R},(4.8/+2) \par \textbf{D},(4.6/-2) \par \textbf{T},(4.4/-0.3) & \textbf{IH},(7/+2.1) \par \textbf{T},(5.5/+1.1) \par \textbf{D},(5.4/-3) \par \textbf{L},(4.7/-2.7) \par \textbf{R},(4.5/-3.8) \\
& \texttt{audio+acoustic} & \textbf{DH},(4.7/-11.6) \par \textbf{S},(4.7/+4.7) \par \textbf{IH},(4.5/+4.5) \par \textbf{AH},(4.5/+2.4) \par \textbf{T},(4.2/+0.2) & \textbf{R},(4.9/-4.1) \par \textbf{S},(4.7/+2.1) \par \textbf{AH},(4.7/+0.8) \par \textbf{T},(4.7/-5.6) \par \textbf{IH},(4.5/+0.6) & \textbf{DH},(5/-14.4) \par \textbf{IH},(5/+5) \par \textbf{T},(5/-1.5) \par \textbf{AH},(4.8/-3.3) \par \textbf{N},(4.4/+4.4) & \textbf{IH},(5.9/-3.2) \par \textbf{N},(4.9/-2.6) \par \textbf{DH},(4.9/-7.6) \par \textbf{D},(4.6/+1.3) \par \textbf{AH},(4.6/-1.2) & \textbf{AH},(5/-0.2) \par \textbf{DH},(4.8/-3.6) \par \textbf{IH},(4.8/-7.8) \par \textbf{T},(4.6/-1.7) \par \textbf{S},(4.5/+4.5) & \textbf{IH},(5.5/+0.2) \par \textbf{AH},(5.3/-1.8) \par \textbf{DH},(5.1/-3.7) \par \textbf{T},(4.8/+1.3) \par \textbf{R},(4.5/-5.6) \\
\midrule
\textbf{Qwen} & \texttt{baseline} & \textbf{DH},(30.6/+19.6) \par \textbf{TH},(21/+20.1) \par \textbf{R},(20.2/+15.9) \par \textbf{V},(6.5/-1.2) \par \textbf{/TH/},(4/+4) & \textbf{DH},(31.4/+24.5) \par \textbf{R},(27.3/+16) \par \textbf{TH},(20.8/+19.9) \par \textbf{L},(3.3/-0.2) \par \textbf{AE},(2/+2) & \textbf{R},(27.5/+20) \par \textbf{DH},(26/+5.2) \par \textbf{TH},(13.6/+10.7) \par \textbf{L},(8.9/+8) \par \textbf{V},(3.9/+1) & \textbf{TH},(26/+22) \par \textbf{DH},(23.5/+10.5) \par \textbf{R},(21.5/+17.5) \par \textbf{/TH/},(4.5/+4.5) \par \textbf{L},(4.2/-3.8) & \textbf{DH},(30.9/+21.3) \par \textbf{TH},(18.9/+15) \par \textbf{R},(7.4/+5.7) \par \textbf{/TH/},(4.5/+4.5) \par \textbf{IH},(4.5/-6.8) & \textbf{R},(17.5/+8.8) \par \textbf{DH},(16.8/+8) \par \textbf{TH},(14.6/+13.9) \par \textbf{/TH/},(9.5/+9.5) \par \textbf{/r/},(6.6/+6.6) \\
& \texttt{ipa} & \textbf{DH},(26.2/+15.2) \par \textbf{TH},(18.8/+17.9) \par \textbf{R},(16/+11.7) \par \textbf{V},(9.2/+1.6) \par \textbf{W},(2.1/+1.3) & \textbf{DH},(24.3/+17.4) \par \textbf{R},(22.8/+11.5) \par \textbf{TH},(14.7/+13.8) \par \textbf{V},(5/+5) \par \textbf{L},(4.6/+1.2) & \textbf{R},(23.5/+16) \par \textbf{DH},(18.5/-2.2) \par \textbf{TH},(13.9/+11.1) \par \textbf{L},(10.9/+10) \par \textbf{V},(4.6/+1.8) & \textbf{TH},(28.4/+24.4) \par \textbf{R},(25/+21) \par \textbf{DH},(23.1/+10.1) \par \textbf{V},(5.2/+1.2) \par \textbf{L},(2.6/-5.3) & \textbf{DH},(20.4/+10.8) \par \textbf{TH},(17.9/+14) \par \textbf{R},(8.2/+6.5) \par \textbf{S},(3.6/+2.5) \par \textbf{HH},(3.2/+2.1) & \textbf{TH},(17/+16.3) \par \textbf{R},(17/+8.2) \par \textbf{DH},(14.4/+5.6) \par \textbf{V},(5.2/+1.6) \par \textbf{W},(5.2/+5.2) \\
& \texttt{acoustic} & \textbf{DH},(25.6/+14.3) \par \textbf{TH},(19.2/+17.3) \par \textbf{R},(18.6/+14.8) \par \textbf{/r/},(3.5/+3.5) \par \textbf{ER},(2.9/-0.9) & \textbf{DH},(23.5/+17.6) \par \textbf{R},(21.4/+7.7) \par \textbf{TH},(12.8/+12.8) \par \textbf{L},(4.3/+0.4) \par \textbf{AH},(2.7/-3.2) & \textbf{R},(24.6/+17.2) \par \textbf{DH},(18.7/-1.9) \par \textbf{TH},(14/+11.1) \par \textbf{L},(7/+5.5) \par \textbf{ER},(5.8/+1.4) & \textbf{R},(25.6/+20.3) \par \textbf{DH},(22.2/+9.1) \par \textbf{TH},(20.6/+17.9) \par \textbf{/TH/},(5/+5) \par \textbf{/DH/},(3.3/+3.3) & \textbf{DH},(24.8/+16.8) \par \textbf{TH},(20/+15.6) \par \textbf{R},(13.9/+11.7) \par \textbf{ER},(4.2/+3.5) \par \textbf{/DH/},(3/+3) & \textbf{DH},(21/+12.4) \par \textbf{R},(18.8/+10.9) \par \textbf{TH},(13.3/+11.1) \par \textbf{W},(4.4/+4.4) \par \textbf{ER},(3.3/-6) \\
& \texttt{audio-only} & \textbf{DH},(28.3/+17.3) \par \textbf{TH},(18.7/+17.9) \par \textbf{R},(12.4/+8.1) \par \textbf{V},(6/-1.7) \par \textbf{/TH/},(2.8/+2.8) & \textbf{DH},(28.3/+21.4) \par \textbf{R},(22.7/+11.4) \par \textbf{TH},(13.4/+12.5) \par \textbf{W},(2.8/+0.2) \par \textbf{V},(2.4/+2.4) & \textbf{DH},(28.4/+7.6) \par \textbf{R},(26.8/+19.3) \par \textbf{TH},(15.2/+12.3) \par \textbf{L},(5.4/+4.5) \par \textbf{V},(3.1/+0.3) & \textbf{DH},(31.8/+18.8) \par \textbf{TH},(30.6/+26.6) \par \textbf{R},(20.2/+16.3) \par \textbf{V},(2.5/-1.5) \par \textbf{L},(2.1/-5.9) & \textbf{DH},(29.1/+19.8) \par \textbf{TH},(23.8/+19.9) \par \textbf{R},(12.3/+10.1) \par \textbf{V},(3.4/-0.4) \par \textbf{IH},(3.1/-8.5) & \textbf{DH},(17.8/+9.1) \par \textbf{R},(16.6/+7.8) \par \textbf{TH},(14.6/+13.9) \par \textbf{/TH/},(3.2/+3.2) \par \textbf{W},(3.2/+3.2) \\
& \texttt{audio+acoustic} & \textbf{DH},(27.2/+16.5) \par \textbf{TH},(19.4/+18.4) \par \textbf{R},(16.7/+13.4) \par \textbf{ER},(2.8/-0.4) \par \textbf{S},(2.8/+2.8) & \textbf{DH},(33.8/+26.9) \par \textbf{R},(24.1/+17.2) \par \textbf{TH},(11/+9.3) \par \textbf{L},(4.8/-0.3) \par \textbf{ER},(2.8/-4.1) & \textbf{R},(33.1/+24.2) \par \textbf{DH},(26.8/+6.8) \par \textbf{TH},(15/+15) \par \textbf{L},(6.3/+4.1) \par \textbf{JH},(3.1/+3.1) & \textbf{TH},(28.9/+24.9) \par \textbf{DH},(25.7/+13.4) \par \textbf{R},(20.4/+16.3) \par \textbf{ER},(3.9/-0.1) \par \textbf{AA},(1.3/-1.7) & \textbf{DH},(25.6/+16.3) \par \textbf{R},(19.3/+17.5) \par \textbf{TH},(18.8/+15.6) \par \textbf{L},(3.4/+2.8) \par \textbf{H},(3.4/+3.4) & \textbf{DH},(20/+10.4) \par \textbf{R},(16.9/+9) \par \textbf{TH},(11.8/+10.1) \par \textbf{W},(3.6/+3.6) \par \textbf{AO},(3.6/+2.5) \\

\bottomrule
\end{tabular}

\label{tab:appendix_phoneme}
\end{table*}

\end{document}